\documentclass[letterpaper, 10 pt, conference]{ieeeconf}  

\IEEEoverridecommandlockouts                              

\usepackage{cite}
\usepackage{amsmath,amssymb,amsfonts}
\usepackage{algorithmic}
\usepackage{graphicx}
\usepackage{textcomp}
\usepackage{lipsum}  
\usepackage{xcolor}
\usepackage[font=footnotesize]{caption}
\usepackage{subcaption}

\begin{document}

\title{\LARGE \bf Precision fruit tree pruning using a learned hybrid vision/interaction controller\\
\thanks{This work has been submitted to the IEEE for possible publication. Copyright may be transferred without notice, after which this version may no longer be accessible.}
\thanks{This research is supported by the United States Department of Agriculture-National Institute of Food and Agriculture through the Agriculture and Food Research Initiative, Agricultural Engineering Program (Award No. 2020-67021-31958).}
\thanks{$^1$Collaborative Robotics and Intelligent Systems (CoRIS) Institute, Oregon State University, Corvallis OR 97331, USA {\tt\footnotesize \{youa, kolanoh, parayiln, cindy.grimm, joseph.davidson\}@oregonstate.edu}}%
\author{Alexander You$^1$, Hannah Kolano$^1$, Nidhi Parayil$^1$, Cindy Grimm$^1$, Joseph R. Davidson$^1$ }
}%


\maketitle

\begin{abstract}
Robotic tree pruning requires highly precise manipulator control in order to accurately align a cutting implement with the desired pruning point at the correct angle. Simultaneously, the robot must avoid applying excessive force to rigid parts of the environment such as trees, support posts, and wires. In this paper, we propose a hybrid control system that uses a learned vision-based controller to initially align the cutter with the desired pruning point, taking in images of the environment and outputting control actions. This controller is trained entirely in simulation, but transfers easily to real trees via a neural network which transforms raw images into a simplified, segmented representation. Once contact is established, the system hands over control to an interaction controller that guides the cutter pivot point to the branch while minimizing interaction forces. With this simple, yet novel, approach we demonstrate an improvement of over 30 percentage points in accuracy over a baseline controller that uses camera depth data.
\end{abstract}

\section{Introduction}
Pruning in the dormant season is an important component of producing high-value, fresh market tree fruit. Pruning involves selectively removing branches that are closely spaced, too old/long, diseased, or otherwise unproductive. This process rejuvenates the tree, helps maintain tree health, and provides high yields of quality fruit at harvest. However, selective pruning is a costly and labor-intensive process, comprising the second highest source of labor costs (after harvesting) for fruits such as apples~\cite{pennsylvania2015penn,gallardo2009cost}. It is also a dangerous task, with workers standing on ladders on uneven terrain with sharp shears. Moreover, the industry at large is facing uncertainty over labor availability in the future~\cite{calvin2010us}. As such, the tree fruit industry is interested in using robots to automate precision pruning operations~\cite{Verbiest2021}.

Automated pruning requires i) sensing and modeling a tree in an outdoor environment; ii) autonomously determining the pruning points (i.e. which branches to remove); iii) efficiently planning a collision-free sequence of cuts; and (iv) executing those precise cuts. The orchard environment is highly unstructured and cluttered, with branches of varying compliance interspersed with rigid features such as trunks, posts, and trellis wires. Navigating pruners --- here we assume that the cutting tool is a set of bypass pruners --- to the desired cut points requires robust collision avoidance, managing the dynamic interaction between the robot and branch, and the ability to locally adapt if the tree shifts or moves. A robotic pruning system must also be robust to missed features in the model (e.g., small branches or wires that are difficult to detect from a distance) and to errors arising from sensor noise and imperfect kinematic models. Furthermore, outdoor robots operate on rough, unstable terrain which can exacerbate errors. All of these errors can lead to scenarios similar to ones we experienced during our previous work~\cite{you2020efficient} where our position-based controller, which relied on depth data with no interaction control regulating contact, nearly snapped a post off of our experimental setup. 

\begin{figure}[bt]
	\centering
\includegraphics[width = \columnwidth]{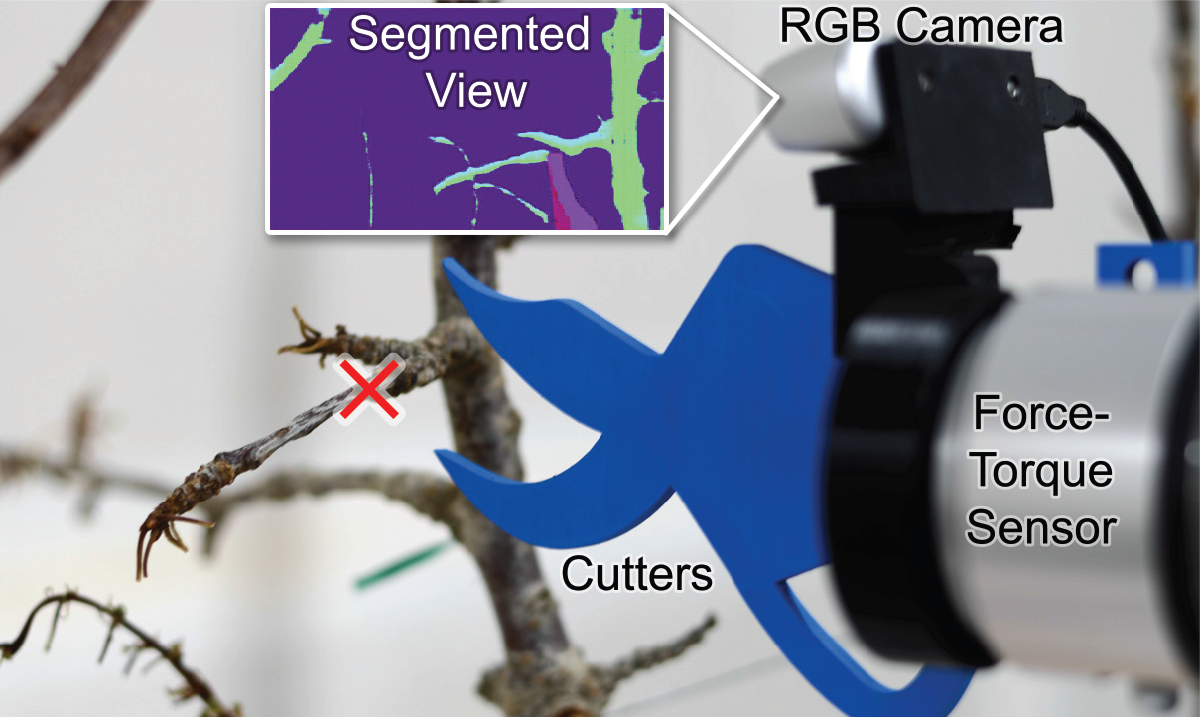}\caption{Our pruning pipeline uses a segmented 2D image along with force measurements for robust execution of precision pruning.}\label{fig:abstract}
    \vspace{-2ex}
\end{figure}

To address these problems, in this paper we propose a hybrid controller (Figure~\ref{fig:abstract}) which splits the pruning task into two stages. First, we utilize a \emph{learned} visual controller which eschews the need for accurate point clouds and kinematics by relying solely on a 2D image to guide the cutters to a desired target, taking into account the correct length of the pruning cut (i.e. the robot cuts the branch where the branch \emph{should} be pruned). This controller handles alignment of the branch and cutter in the vertical and horizontal directions of the image plane (perpendicular to the cutter). Since a visual controller alone cannot handle delicate environmental interactions, once we detect contact with the tree, we transition to an admittance controller incorporating force feedback. This controller has two main tasks: limiting the forces between the robot and the environment, and accurately guiding the branch towards the pivot point of the cutter for an optimal, precise cut.  

We evaluate our system on a test setup using branches obtained from a cherry tree fixed to a rigid trellis. We compare our hybrid controller against a set of position-based controllers which rely directly on a point estimate produced from a depth camera point cloud. We demonstrate an improvement in cutting accuracy of over 30 percentage points compared to the best position-based controller. Furthermore, our system is able to more consistently cut branches at a pre-specified length, cutting the standard deviation of such cut lengths nearly in half compared to the baseline controllers.

\section{Related work}

A comprehensive review of robotic pruning is given in~\cite{zahid2021}, focusing on subproblems such as branch sensing, pruning point detection, end-effector design, and path planning. While a fairly substantial amount of work has gone into these subproblems, research on controllers that bring a physical end-effector to pruning points is sparse. To our knowledge, the only work using a physical manipulator to reach pruning points was our own~\cite{you2020efficient} and those from~\cite{visrob} and~\cite{botterill2017robot} (grapevines). Zahid et al.~\cite{zahid2020collision} presented a pipeline for collision-free planning for apple tree pruning, but the work was evaluated only in simulation. Because of the lack of work on physical pruning systems, we focus on hybrid force controllers and reinforcement learning schemes for robotic manipulation in general.



Hybrid force-control systems have been developed for a variety of manipulation problems where both visual and force feedback are required, such as door opening~\cite{prats2009vision, Schmid2008}, peg-in-the-hole assembly problems~\cite{mezouar2007external}, and surface-based point/contour tracking applications~\cite{leite2006hybrid,Baeten2002}. A  common control method is to partition the degrees of freedom of control between the visual and force components. This method is similar to our own in which the vision controller handles vertical-horizontal alignment while the interaction controller handles vertical-forward alignment (though we choose to run our two controllers separately). However, these methods were often limited by the computer vision techniques employed in the visual controllers, which typically assumed highly predefined environments with known markers or models for easy target detection and pose estimation.

Modern deep learning-based, image-based approaches to grasping and manipulation have shown success in generalizing across large object sets, allowing for more robust manipulation. A comprehensive review of deep learning methodologies and applications for robotic manipulation can be found in~\cite{Mohammed2020}. While such algorithms have obtained success in moving the object to be manipulated inside of the end-effector, many of the vision-only controllers assume that compliance in the manipulator will prevent the end-effector/manipulator or object from being damaged by the interaction. In contrast, pruners are very rigid, sharp objects interacting in an environment with both delicate, compliant objects (branches, buds, leaves) and stiff objects (wires, posts, large branches) which can damage the cutter. This motivates the need to utilize force feedback once contact is made instead of continuing with a vision-only approach.

\section{System design}
\label{sec:overview}

\begin{figure}[bt]
	\centering
\includegraphics[width = 0.9\columnwidth]{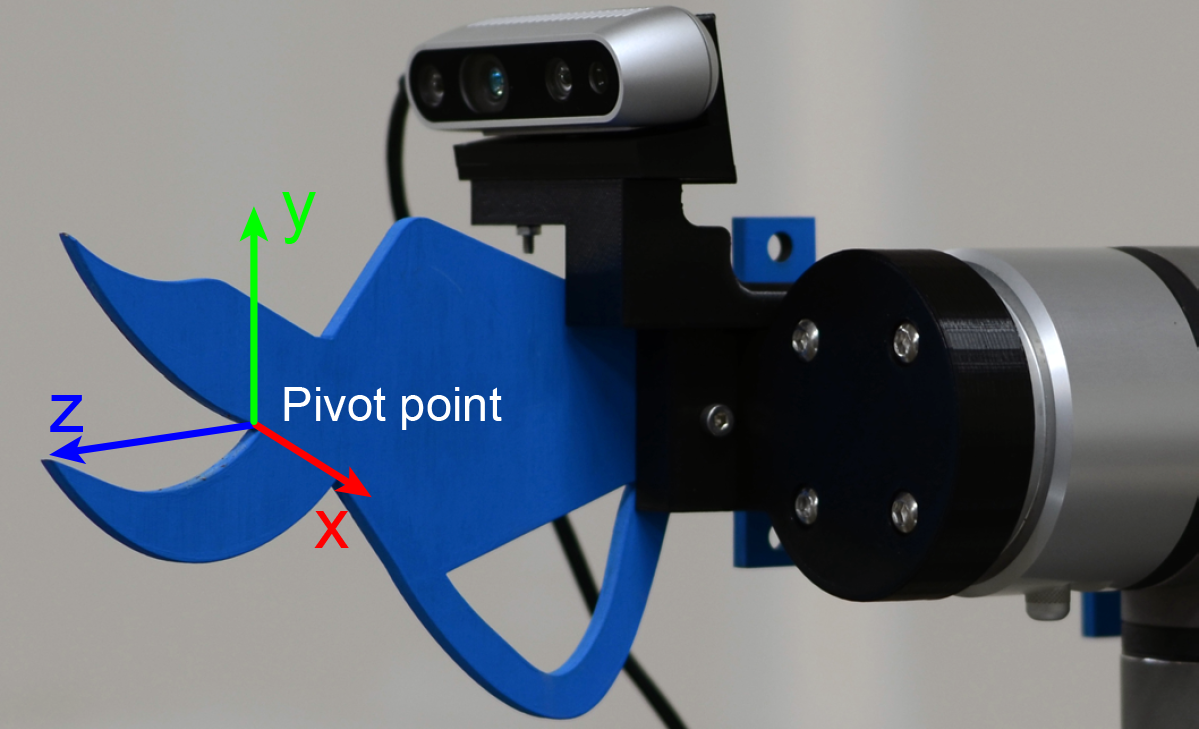}\caption{Our mock pruning end-effector with a Realsense D435 camera mounted on the wrist. The tool reference frame is also shown, where the $x$-axis is left-right, the $y$-axis is up-down, and the $z$-axis is in-out. The camera is placed so that the top of the cutter and the target branch are visible.}\label{fig:robotsetup}
    \vspace{-2ex}
\end{figure}

For our tree pruning system design, we make the assumption that we have a 3D tree model with predetermined pruning points, and that these pruning points, when converted to the robot's base frame, are roughly aligned with the corresponding points on the real tree. Furthermore, we assume that there are no notable obstructions in front of any targets that would completely obstruct a cutting implement from moving towards the goal. Finally, we assume that we are working within a planar tree environment, similar to modern high-density orchard systems, which means that all targets should be discernible from a 2D image taken with a camera pointed towards the tree. This assumption, plus the general noisiness of depth data from depth cameras, motivated us to build a control system that uses 2D image data to align the tool with the target branch.

Our robot system is shown in Figure~\ref{fig:robotsetup}. It consists of a Universal Robots UR5e (Odense, Denmark) six degree-of-freedom industrial manipulator with a replica 3D pruning implement attached to the robot's tool flange. We mount an Intel RealSense D435 (Santa Clara, CA, USA) on top of the cutter such that the cutting implement is in the camera's field of view, allowing for accurate vertical and horizontal positioning of the pruners with respect to the tree. The goal is to navigate the cutters so that the branch is located at the pivot point of the cutters, minimizing the amount of force required to prune the enclosed branch~\cite{zahid2021evaluation}.

\begin{figure*}[bt]
	\centering
\includegraphics[width = \linewidth]{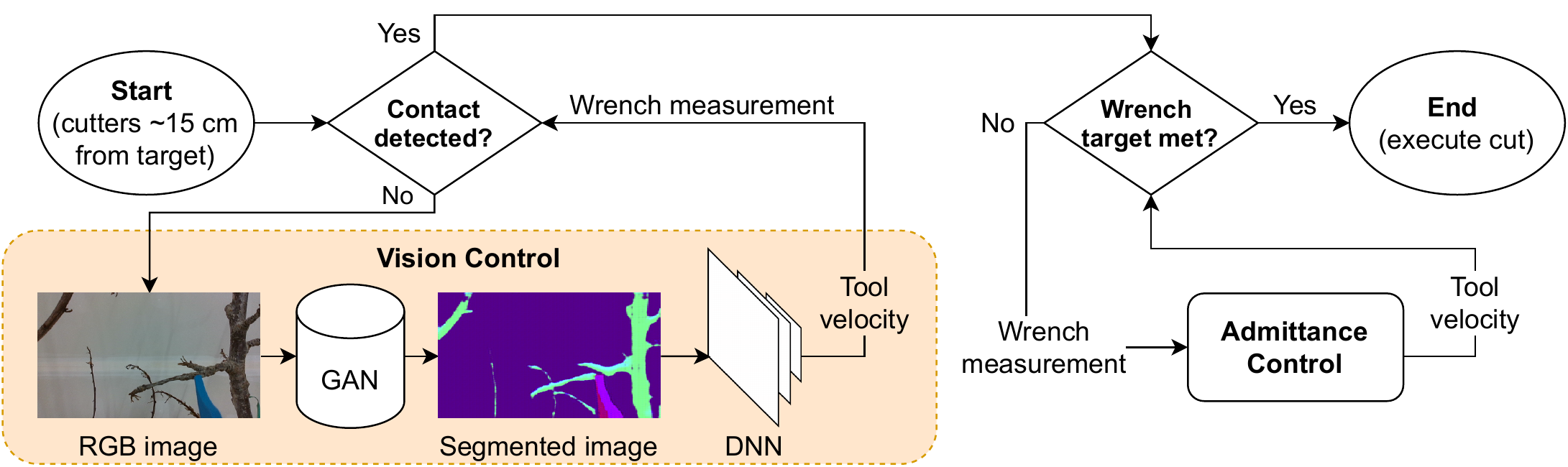}\caption{An overview of our precision pruning system. Initially, we use a vision-based controller that processes RGB images of the environment into a segmented image and feeds that image through a neural network to produce a velocity command. Upon contact, we switch to an admittance controller to guide the branch into the cutter pivot point for optimal cutting.}\label{fig:systemchart}
    \vspace{-2ex}
\end{figure*}

The control flow of our pruning process is shown in Figure~\ref{fig:systemchart}. Our hybrid control system is divided into two main systems: A vision-based controller which focuses on moving the cutter to the general proximity of the target branch (Section \ref{sec:vision}), followed by a force-based admittance controller which aligns the branch with the cutter pivot point (Section \ref{sec:force}). 

\section{Vision system}
\label{sec:vision}


The goal of the vision system is to guide the cutters so that the branch is enclosed in the cutter jaws. We assume the robot end-effector is initially positioned 15-20 cm from the target branch, the camera sees both the cutter and the target branch, and the branch orientation is roughly orthogonal to the cutter mouth. 

Our primary challenge is building a controller that is easily trainable and robust to a wide variety of lighting, branch shape, color, texture, and background complexities. To do this, we adopt the approach in~\cite{James2019} and train a Generative Adversarial Network (GAN)~\cite{goodfellow2014generative}, which essentially segments out the branches in the scene. Importantly, this GAN performs similarly for both simulated and real images, allowing us to train our controller entirely in simulation. We then train a deep neural network using an actor-critic reinforcement learning paradigm. The neural network takes in a segmented image and outputs a vertical and horizontal control action to move the robot, with the goal of positioning the branch inside the cutter jaws.


First, we cover the simulation environment used to collect simulated data (Section~\ref{sec:simulation}). Next, we discuss how we adapt existing work on sim-to-real transfer the aforementioned GAN (Section~\ref{sec:realtosim}). Finally, we formally define the reinforcement learning problem (Section~\ref{sec:rl}).



\subsection{Simulation environment}
\label{sec:simulation}

\begin{figure}[bt]
	\centering
\includegraphics[width = \columnwidth]{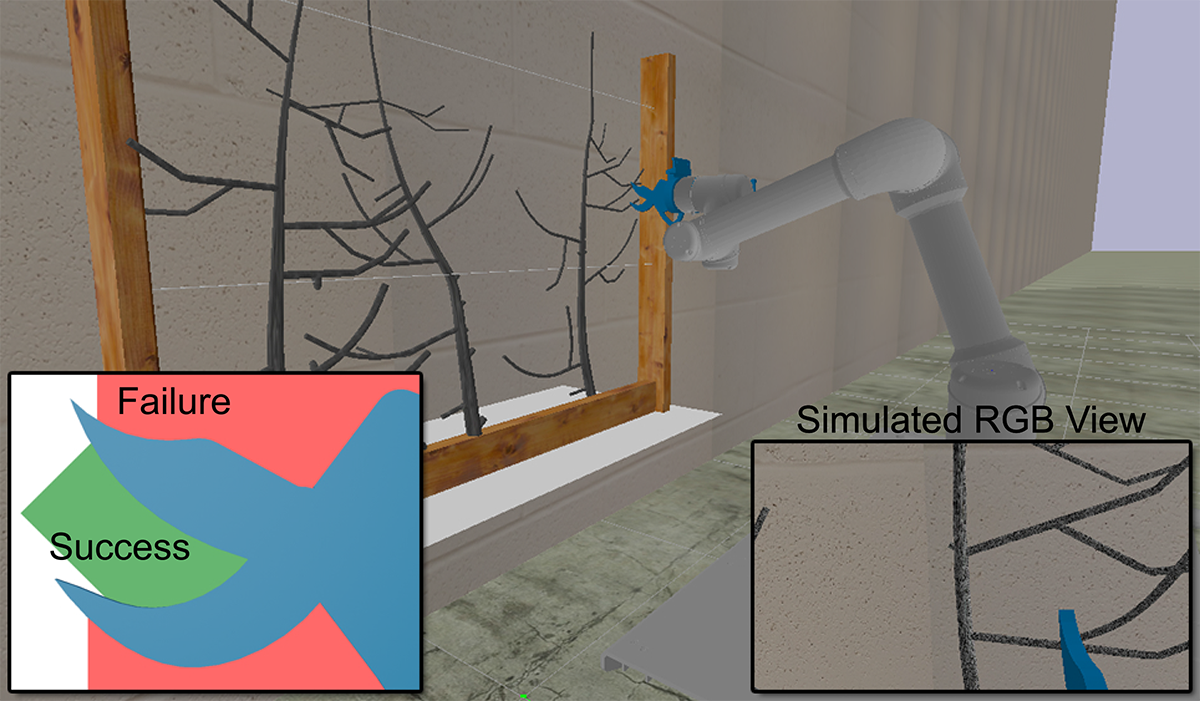}\caption{Our simulation environment, constructed in PyBullet, along with the success and failure regions for the cutter. These are used to query if the branch has entered (or missed) the cutter jaws.}\label{fig:sim}
    \vspace{-2ex}
\end{figure}

We mimic our physical experimental setup (Section~\ref{sec:expsetup}) in the PyBullet physics engine~\cite{coumans2016pybullet} (Figure~\ref{fig:sim}). The setup consists of the UR5e arm mounted on a cart with the cutter as an end-effector, a trellis system consisting of a wooden frame, wires, and three spindle tree models randomly chosen from our 8 pre-generated models. Each of the tree spindles is annotated with several pruning targets on the side branches, located approximately 3 cm from the upright leaders. The tree models are assumed to be completely stiff.

We render RGB images of the simulated environment from the virtual camera to use as input to the GAN. We also compute the distance from the cutter to the given target cut point and use collision detection to query if the target branch has entered the cutter or not (Figure~\ref{fig:sim}). Since the trees are completely stiff, the collision physics between the robot and the environment are not realistic. Therefore, we extend the success region slightly beyond the cutter mouth. The assumption is that subsequent movements of the arm will not cause the branch to leave the success region.

\subsection{Real-to-sim transfer}
\label{sec:realtosim}


Our approach to the sim-to-real transfer follows the approach of \cite{James2019}, which trains a GAN to transform the real image into a ``synthetic'' version --- essentially performing a segmentation. They train the GAN with simulated images of the environment generated using a wide variety of lighting, colors, and textures. This heavy randomization allows the system to generalize to real images surprisingly well. The simple, segmented image essentially decouples the visual processing of a complex scene from the decision of what action to take given where the branch is.

We follow their methodology to create a 3-channel segmented representation of our environment. First, we identify five unique elements in the scene: The cutter, the trees, the wooden trellis frame, the trellis wires, and everything else (i.e. the walls and floor). We replace the object's true texture with a unique uniform color texture for each element and render the scene again with the simplified colors. Next, we convert the RGB image into the HSV (hue-saturation-value) space. Finally, we replace the S channel with a segmentation mask from PyBullet specifically for the trees (255 if the pixel corresponds to a tree and 0 otherwise).

    

\begin{figure}[bt]
	\centering
\includegraphics[width = \columnwidth]{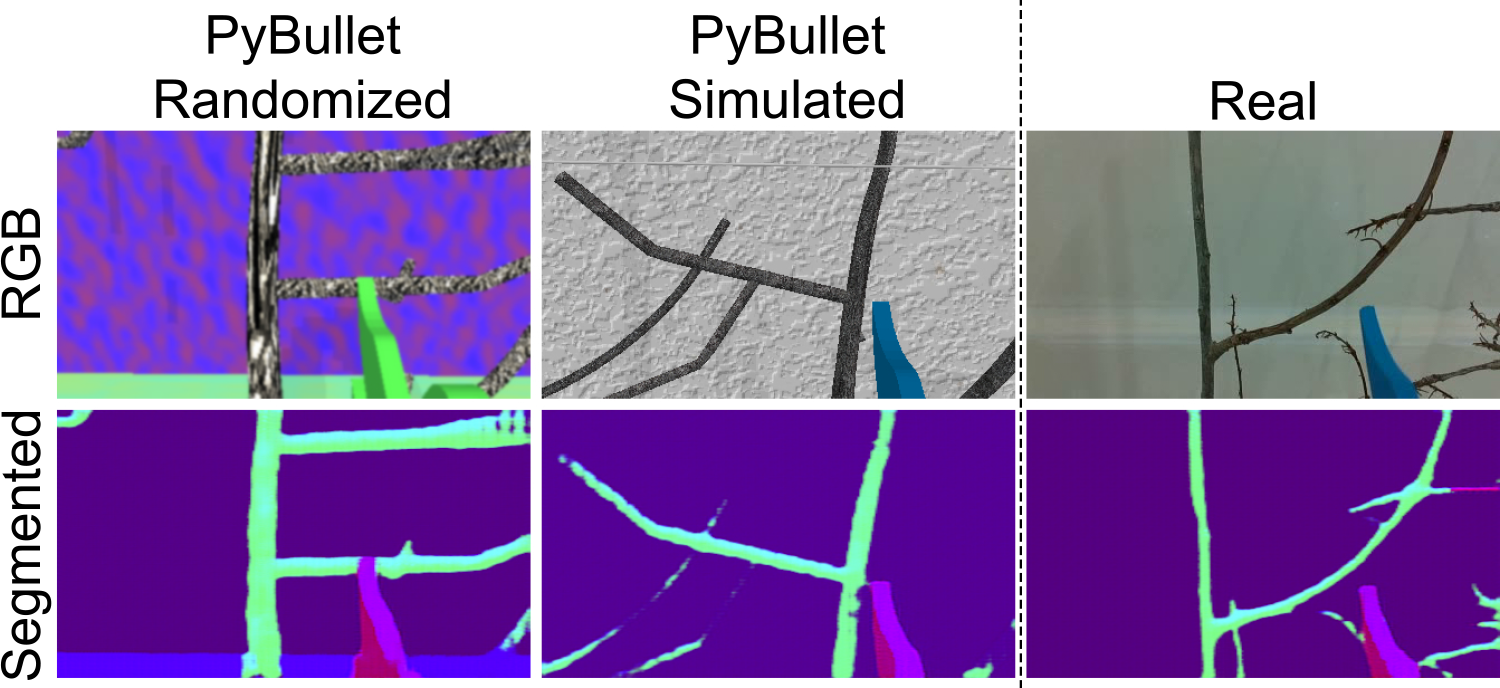}\caption{Examples of our GAN converting images of different types, including simulated and real images, into their segmented form.}\label{fig:ganex}
    \vspace{-2ex}
\end{figure}

To train the GAN, we pair each segmented environment representation with a corresponding randomized version in which each object is assigned one of 5000 random synthetic textures and the lighting's source, color, and intensity have been randomized. We generate 10,000 randomized-segmented image pairings and utilize the pix2pix architecture~\cite{pix2pix2017} to learn how to translate the randomized images to their segmented form. Examples of the trained GAN working on different images are shown in Figure~\ref{fig:ganex}. 


\subsection{Reinforcement learning}
\label{sec:rl}





We now formally define a pruning episode, as well as the state space $\mathcal{S}$, the action space $\mathcal{A}$, the reward function $\mathcal{R}$, and the transition function $\mathcal{T}$ which define a Markov Decision Process (MDP). 
Each episode is given a maximum time length $T$ s with a time interval of $\delta_t$ s, resulting in a maximum number of timesteps of $N_t = \frac{T}{\delta_t}$. At the start, we select a target cut $p_{cut}$ on a random spindle and position the cutter 15-20 cm in front of $p_{cut}$.

At each timestep we obtain a 424 x 240 segmented image from the GAN. Since the cutter and branch are in the right of the image, we crop the bottom-right 360 x 180 of the image and rescale it to 160 x 80. Therefore, our state space is the set of 160 x 80 3-channel images ($\mathcal{S} = Im_{160 \times 80 \times 3}$). 

The segmented image is fed through a deep neural network which produces two numeric values, $a_x, a_y \in [-1,1]$, representing the velocity outputs ($\mathcal{A} = [-1, 1] \times [-1, 1])$. By default, the cutter moves in the tool frame's $z$-axis at a speed of $s$ m/s. The chosen action values modify the $x$ and $y$ components of this velocity vector, such that the final velocity vector is $\overrightarrow{a} = \langle a_x s, a_y s, s \rangle$. The transition function $\mathcal{T}$ maps to the next state by setting the tool velocity to $\overrightarrow{a}$ and running the simulator forward for $\delta_t$s. 

The reward function $\mathcal{R}$ is engineered to reward the cutter for approaching the target cut point, giving more reward as the cutter approaches the target, while heavily penalizing attempts which end up missing the cutters. The total reward per episode ranges between $-T$ and $T$. Each timestep's reward depends only on the state $s_{n+1}$ reached after applying action $a_n$ at state $s_n$. At each new state $s_{n+1}$, we check for one of three terminal conditions:

\begin{itemize}
    \item (Success) The branch has entered the success zone.
    \item (Failure) The branch has entered the failure zone.
    \item (Failure) The episode has reached $N_t$ time steps.
\end{itemize}

Upon success, the agent receives a reward corresponding to the remaining time $T - \delta_t n$, while upon failure, it receives a large penalty of $-T$. Otherwise, the reward is a value between $0$ and $\delta_t$ depending on how close the cutter is to the target and a minimum distance threshold $d_{min} = 0.10$ m:

\begin{equation}
    \mathcal{R}(s_{t+1}) = \delta_t \max{\left(1 - \frac{d_{t+1}}{d_{min}} , 0\right)}
\end{equation}

For training in simulation, we set $T = 1$ s and $\delta_t = 0.10$ s ($N_t = 10$). To encourage fast episode completion, we set the forward velocity equal to $s = 0.30$ m/s. For the real robot arm, we limited the forward velocity to $s = 0.03$ m/s.

We use the Proximal Policy Optimization (PPO) algorithm~\cite{schulman2017proximal} from the \texttt{stable-baselines3} package~\cite{stable-baselines3} to learn the desired control actions. We use the default hyperparameters and deep neural network architecture. This consists of a feature extractor with three convolutional layers and one feedforward layer with ReLU activation. The action and value networks share the output of the feature extractor and process the features through their own feedforward layers to obtain the velocity command and value estimates.



\section{Admittance controller}
\label{sec:force}

\begin{figure}
    \centering
    \begin{subfigure}[b]{0.40\columnwidth}
        \centering
        \includegraphics[width=\columnwidth]{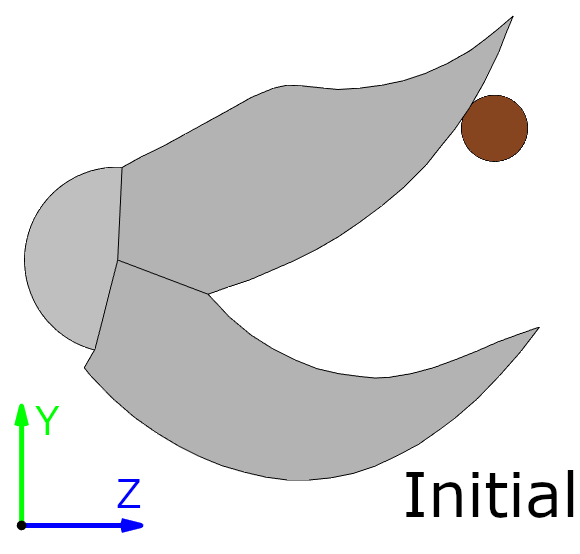}
        \label{fig:sub-contact}
    \end{subfigure}
    \hfill
    \begin{subfigure}[b]{0.40\columnwidth}
        \centering
        \includegraphics[width=\columnwidth]{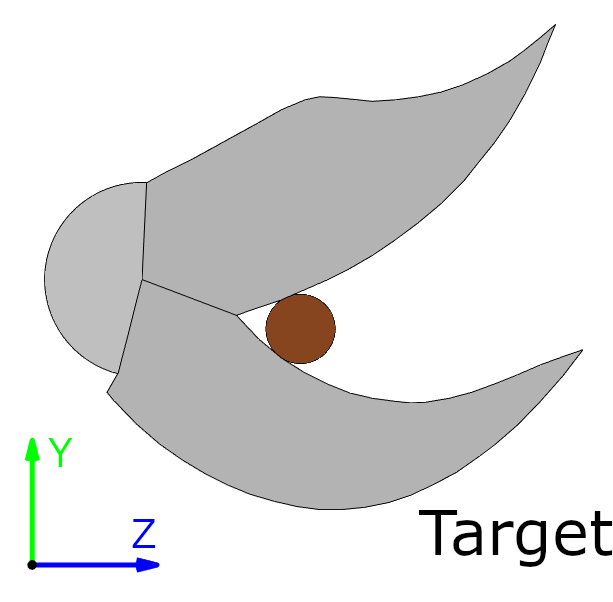}
        \label{fig:sub-pivot}
    \end{subfigure}
    \vspace{-2ex}
    \caption{The admittance controller guides the branch from initial contact to the pivot point. Left: A possible contact point. Right: The branch at the pivot point, ideal for executing a cut.}
    \label{fig:matlab}
\end{figure}

Once a force exceeding $0.75$ N is detected, we assume that the target branch has contacted a cutting surface and transition to an interaction controller. Regulating the dynamic interaction with the branch is critical for two reasons. First, we want to minimize contact forces to prevent damage to the tool/robot. Second, to deliver maximum cutting force, the branch should be located as close to the blade pivot point as possible (see Figure~\ref{fig:matlab}). 

The contact task is characterized by interaction between a `stiff' industrial manipulator coupled with an elastically compliant branch. We choose to use admittance control: the robot senses external forces using its internal wrist force-torque sensor and controls its motion in response. An admittance controller imitates the dynamics of a spring-mass damper in the robot's task space, as described by Equation~\ref{eq:dynamics}:

\begin{equation}
    M \ddot{x}_c + B \dot{x} + K x = \mathcal{F}_{ext}
    \label{eq:dynamics}
\end{equation}

Here, $M$, $B$, and $K$ are the mass, damping, and stiffness matrices to be imitated, $(x, \dot{x})$ are the end-effector position and velocity in task space, $\mathcal{F}_{ext}$ is the wrench exerted on the environment, and $\ddot{x}_c$ is the output acceleration as determined by the controller. Since is there no desired ``position'' of the end-effector, we can eliminate the stiffness term and rearrange Equation~\ref{eq:dynamics} as shown in Equation~\ref{eq:dynamicssimplified}:

\begin{equation}
    \ddot{x}_c = M^{-1} (\mathcal{F}_{\mbox{ext}} - B \dot{x})
    \label{eq:dynamicssimplified}
\end{equation}

Once the branch is in contact, we assume that $x$-axis (side to side) motions are unnecessary. This is captured in a square selection matrix $\Lambda$ with a  diagonal of $diag(\Lambda) = \begin{bmatrix} 0 & 0 & 0 & 0 & 1 & 1 \end{bmatrix}$. Multiplying $\Lambda$ by $\mathcal{F}_{\mbox{ext}}$ ensures that the admittance controller only generates motion in the vertical and forward axes. 

Because the force-torque sensor is inherently noisy, we impose both a moving average filter and a deadzone function on the raw wrench data. The sensor publishes at $500$Hz, so we implemented a $51$-value moving average filter to create $\mathcal{F}_{\mbox{ext}}' = \mathrm{filter}(\mathcal{F}_{\mbox{ext}})$. While this results in a $0.1$ s delay, we assume that the maximum speed of the cutter (currently  $1$ cm/s) is small enough to mitigate any significant instability.

We also impose a deadzone function in which all force errors $F_e$ under some threshold $F_{th} = 0.2$ N are set to zero so that small force variations do not produce motion. The deadzone function is defined in Equation~\ref{eq:deadzone}: 

\begin{equation}
    \mathrm{dz}(F_e) = sgn(F_e) \max(\left|F_e\right| - F_{th}, 0)
    \label{eq:deadzone}
\end{equation}



We define the desired end-effector wrench to be $\mathcal{F}_{\mbox{des}}$, and define the desired velocity to be all zeroes. Thus, the full admittance controller is as shown in Equation \ref{eq:admittance}:

\begin{equation}
    \ddot{x}_c = M^{-1} (\mathrm{dz}(\mathcal{F}_{\mbox{des}}-\Lambda\mathcal{F}_{\mbox{ext}}') - B \dot{x})
    \label{eq:admittance}
\end{equation}

For our experiments, we defined the desired wrench to be 
 \begin{equation}
     \mathcal{F}_{\mbox{des}} = \begin{bmatrix} 0 & 0 & 0 & 0 & 0 & -2\end{bmatrix}
 \end{equation}
 
 \noindent in which a $2$ N force is imposed forwards on the branch. We designed the hardware such that the cut point and sensor are located at the same height; therefore, when the branch is at the pivot point, the moment about the $x$ axis will be $0$. The controller declares the objective complete when the sensed torque in the $x$ direction is within $0.0025$ N-m of the desired torque and the tool has moved less than $.5$ mm upward or forward in the last $1$s.
 
 We set the controller gains to diagonal matrices with $diag(M) = \begin{bmatrix} 0 & 0 & 0 & 0 & 100 & 10 \end{bmatrix}$ and $diag(B) = \begin{bmatrix} 0 & 0 & 0 & 0 & 400 & 250 \end{bmatrix}$. On the real robot, controller commands were sent to the UR5e as discrete velocities: $\dot{x}_n = \dot{x}_{n-1} + \ddot{x}_c*dt$.

\section{Experiments}


\label{sec:experiments}

\subsection{Experimental Setup}
\label{sec:expsetup}

\begin{figure}[bt]
	\centering
\includegraphics[width = 0.8\columnwidth]{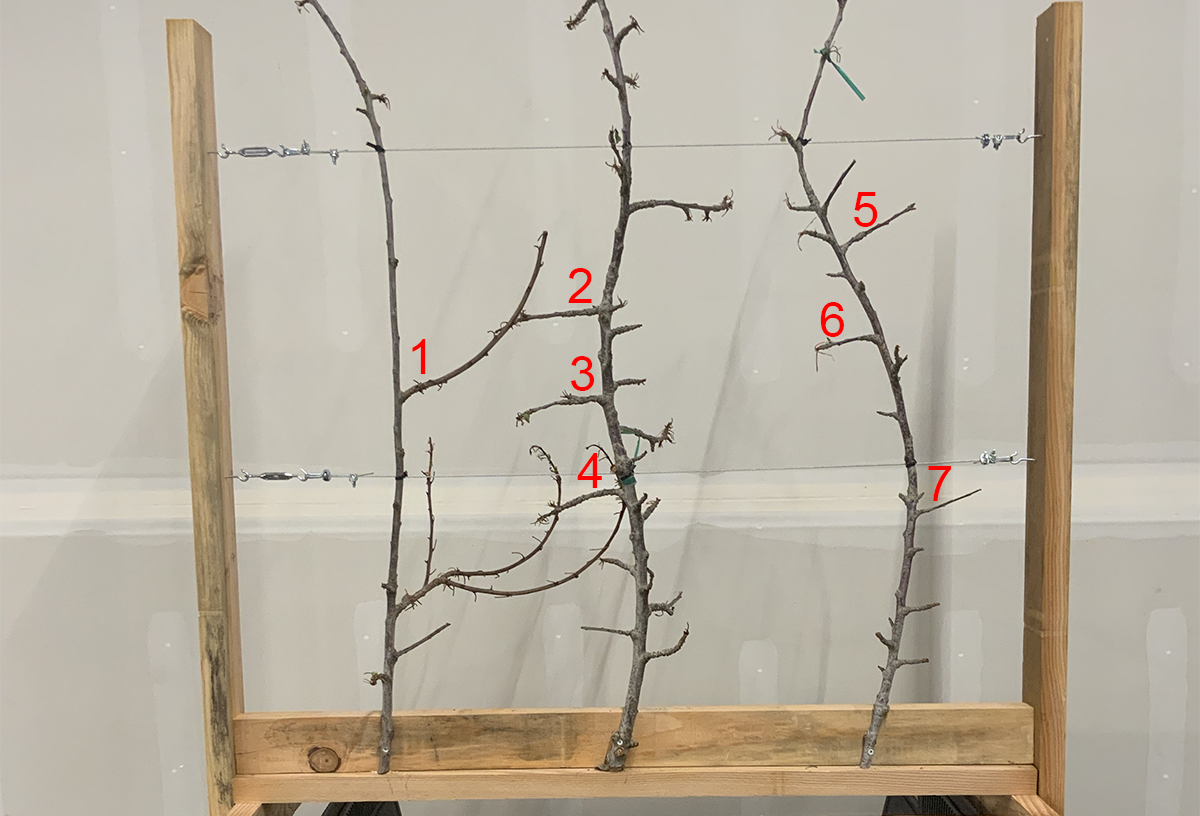}\caption{Our experimental testbed, consisting of several leader branches from a sweet cherry tree affixed to a trellis system.}\label{fig:setup}
    \vspace{-2ex}
\end{figure}

We set up a mock trellis system (Figure~\ref{fig:setup}) to compare our controller with a baseline controller in a physical environment. This setup is based on the structure found in modern orchards, consisting of a wooden frame with two trellis wires running across the frame. We then affix three tree branches, obtained from a real unpruned sweet cherry tree, to the base of the frame and tie the branches down to the trellis wires. This gives the branches compliance properties similar to those found in an actual orchard. We identified seven viable pruning target branches on the leaders (i.e. smaller side branches growing from the main leaders).

\begin{figure*}[bt]
	\centering
\includegraphics[width = 0.9\textwidth]{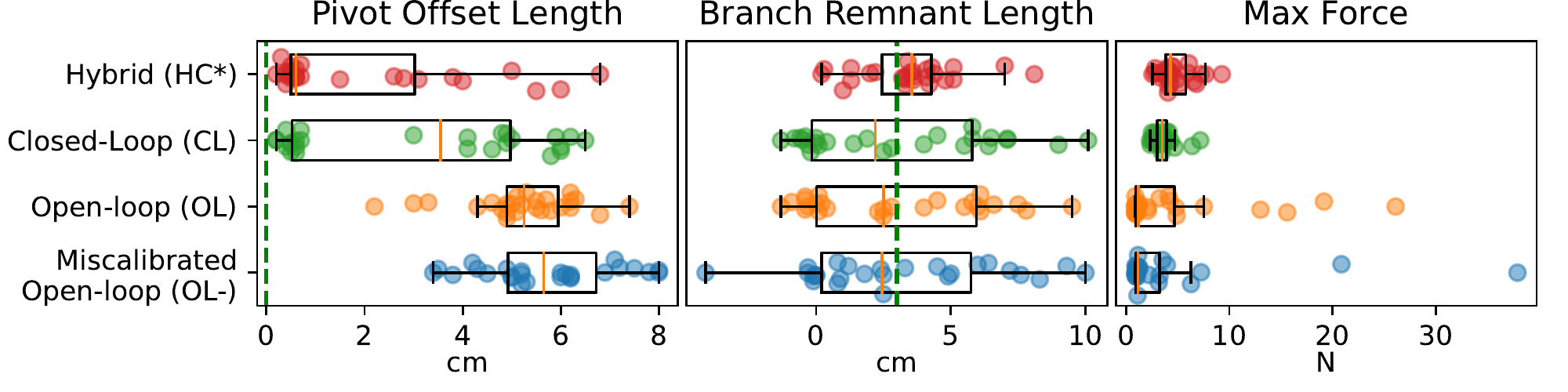}\caption{Results from running each of our controllers for 26 trials. Dashed lines represent ideal targets. Our hybrid controller (HC*) is significantly more accurate and consistent with its cuts than all comparison controllers while successfully avoiding large forces on the environment. (Each box encompasses the lower and upper quartiles, with the whiskers encompassing points within 1.5 interquartile range of the lower/upper quartiles.) The very large forces on the open loop controllers were due to the cutter driving into the vertical leader branches, rather than the side branches.}\label{fig:results}
\end{figure*}

\subsection{Comparison controllers}
\label{sec:controllers}

In order to evaluate our hybrid controller we define three position-based controllers. These were chosen to independently evaluate our controller versus a position-based one, and the effectiveness of the force-based controller over a vision-based one. Specifically:

\begin{itemize}
    \item (OL) A simple open-loop Cartesian-space proportional controller (with maximum velocity of 3 cm/s) which stops when the cutter pivot is within 1 mm of the target.
    \item (CL) A closed-loop controller which sets the target 10 cm past the target estimate and uses the open-loop proportional controller, but transitions to the admittance controller upon contact.  
    \item (OL-) A miscalibrated open-loop position controller. This is the same as the open-loop position controller, but when estimating the target position, the pose of the camera is perturbed by 1 cm in a random direction with a rotation of 5 degrees about a random axis, resulting in a different target estimate than the other methods.
\end{itemize}

\subsection{Experimental trials and evaluation criteria}
\label{sec:trials}

For each trial, we first selected one of the seven target pruning points, each of which was marked with a removable piece of tape 3 cm from the leader. After driving the robot back to one of four home positions, each of which provided a distinct view of the tree (from the front, left, right, and tilted up), we selected a point from the point cloud corresponding to the target, giving us a 3D estimate of the desired pruning cut point (including a perturbed estimate from the miscalibrated controller). We then planned a path so that the cutter was 15 cm away from the estimated target, oriented normal to the trellis plane, and then removed the piece of tape marking the target.

At this point, we run each of the four control systems in succession (our hybrid controller plus the three position-based controllers) and recorded the following metrics to evaluate the performance of each cut:

\begin{itemize}
    \item Accuracy: Is the target branch in the mouth of the cutter?
    \item Pivot offset length: Distance of the branch to the cutter pivot, measured by hand.
    \item Branch remnant length: The length of the branch that would have remained after a cut along the cutter's $yz$ plane (from the plane of the cut surface to the intersection with the vertical leader), measured by hand.
    \item Max force: The maximum force magnitude from the UR5e's force-torque sensor during the execution. 
\end{itemize}

In total, for each of the four controllers, we ran 26 trials, corresponding to 4 trials each for 6 of the targets and 2 trials for one target which broke off during the experiment.


\section{Results}

\begin{table}[]
\begin{tabular}{|c|cccc|}
\hline
 & Acc. & Pivot Len (cm) & Rem Len (cm) & Max Force (N) \\ \hline
HC*         & \textbf{77\%}     & \textbf{1.9 $\pm$ 2.1}   & \textbf{3.5 $\pm$ 1.8}    & 4.8 $\pm$ 1.7  \\
CL         & 46\%     & 3.0 $\pm$ 2.5   & 3.0 $\pm$ 3.5    & \textbf{3.6 $\pm$ 1.1}  \\
OL         & 19\%     & 5.2 $\pm$ 1.1   & 3.0 $\pm$ 3.3    & 4.6 $\pm$ 6.6  \\
OL-        & 23\%     & 5.7 $\pm$ 1.4   & 3.1 $\pm$ 3.6    & 4.1 $\pm$ 8.0  \\\hline
\end{tabular}
\caption{Averaged summary stats of our results for each controller (defined in Section~\ref{sec:controllers}). Standard deviation values are given to the right of the $\pm$ symbol.}
\label{table:results}
\end{table}

A summary of the average results is shown in Table~\ref{table:results}, along with a detailed plot of the values in Figure~\ref{fig:results}. In summary, our hybrid controller performed significantly better than the closed-loop controller in terms of cutting accuracy and precision. Both closed-loop controllers performed better than both of the open loop controllers (which generally failed to complete the task). Our hybrid controller had a 31 percentage point increase in accuracy over the closed-loop controller. Furthermore, it managed to avoid bumping into the upright leaders of the tree on every trial and achieved relatively consistent branch remnant lengths, with a standard deviation in remnant length of just 1.8 cm. This is in comparison to the closed-loop controller, which had nearly double the standard deviation at 3.5 cm and sometimes drove straight into a leader branch. Our controller also more consistently ended up with low pivot offset lengths, though this was largely a result of situations in which the cutter was placed in front of a vertical leader: our controller would move the cutter away from the leader, while the closed-loop controller would move straight into it.


Generally speaking, the depth estimates provided from the camera fell short of the actual targets, which meant that in most of the open-loop runs, the cutter simply stopped in front of the target without ever touching it. However, on the occasions that the open-loop controllers did hit the branch, because of the lack of force feedback, they would push into the branch with a very high amount of force, sometimes upwards of 20 N as a result of hitting a vertical leader branch. Meanwhile, both our hybrid controller and simple closed-loop controller kept all forces below 10 N, demonstrating the value of the interaction controller in preventing damage to the environment. Although our controller did generally incur higher forces than the simple closed-loop controller, likely due to the acceleration of the tool from responding to the neural network, at no point did it ever hit the environment with enough force to cause any notable damage. 

One note is that there appeared to be some error in our ``ground truth" kinematics when transforming point cloud points into the world frame. This was evidenced by two observations: the similarity of the two open loop controllers' overall performance, and the fact that when running the experiments, left-side branch target estimates tended to be far away from the leader while right-side branch target estimates tended to be very close to the leader. While locating and fixing the calibration issues may have improved the overall performance of the simple closed-loop controller, it would also reinforce the point that our system is robust to imperfect kinematic calibration, as well as other perturbations such as cart or tree movement that a point estimate-based controller simply has no way to handle.


\section{Conclusion}

In this paper, we introduced a framework for highly accurate robotic pruning that uses 2D RGB images and force data to consistently execute accurate cuts with minimal force. Through our experiments, we demonstrated that our system is robust to a number of issues that have traditionally been associated with depth camera data, primarily kinematic miscalibrations and incorrect depth estimates. While we conduct our experiments in an indoor laboratory setting, our framework represents a step towards moving robotic precision pruning to the outdoor orchard environment. Future work will include adding rotation as a third dimension to the velocity command (to ensure perpendicular alignment), as well as determining how to adjust the segmentation network to work in more complex outdoor settings with background noise from adjacent rows.


\bibliographystyle{./bibliography/IEEEtran}
\bibliography{./bibliography/pruning2022.bib}

\end{document}